\pdfoutput=1

\documentclass[11pt]{article}

\usepackage[final]{acl}

\usepackage{times}
\usepackage{latexsym}
\usepackage{caption}
\usepackage{multirow}
\usepackage[T1]{fontenc}
\usepackage{enumitem}

\usepackage[utf8]{inputenc}

\usepackage{microtype}

\usepackage{inconsolata}

\usepackage{hyperref}
\usepackage{url}
\usepackage[table]{xcolor} 

\usepackage{subcaption} 
\usepackage{placeins}
\usepackage{wrapfig}
\usepackage{booktabs}
\usepackage{graphicx}
\usepackage[cmex10]{amsmath}
\usepackage{amssymb}
\usepackage{algorithmic}
\usepackage{algorithm}
\usepackage{amsthm}
\usepackage{tcolorbox}

\DeclareMathAlphabet\mathbfcal{OMS}{cmsy}{b}{n}
\newcommand{\ten}[1]{\mathbfcal{#1}}
\newcommand{\mat}[1]{\mathbf{#1}}

\newcommand{\vect}[1]{\boldsymbol{#1}}
\newcommand{\textgray}[1]{\textcolor{gray}{#1}}

\title{LoRi: Low-Rank Distillation for Implicit Reasoning}

\author{
\textbf{Ryan Solgi\textsuperscript{1},  Jiayi Tian\textsuperscript{1},
Zheng Zhang\textsuperscript{1}} \\
\textsuperscript{1}University of California-Santa Barbara, USA \\
\texttt{solgi@ucsb.edu}, \texttt{zhengzhang@ece.ucsb.edu}
}

\begin{document}

\maketitle

\begin{abstract}
Implicit chain-of-thought (iCoT) methods aim to internalize reasoning in large language models, but often underperform explicit CoT prompting. We empirically find that hidden-state reasoning trajectories exhibit low-rank structure. Motivated by this observation, we propose a low-rank distillation framework that transfers reasoning by aligning teacher and student trajectories in a shared low-rank tensor subspace using first- and second-order statistics. The resulting formulation captures the global structure of reasoning while supporting a compact latent reasoning process. We evaluate the method across multiple model families, including LLaMA and Qwen, at different scales on mathematical reasoning benchmarks. Our approach consistently improves performance, especially on challenging multi-step tasks, approaching explicit CoT accuracy and outperforming prior iCoT distillation methods.
\end{abstract}

\section{Introduction}

Large language models (LLMs) exhibit strong reasoning abilities under explicit chain-of-thought (CoT) prompting~\citep{wei2022chain, zelikman2022star}. However, explicit CoT is computationally expensive, can encourage reliance on non-robust patterns, and is sensitive to inference procedures~\citep{li2024chain, lin2025critical, wang2022selfconsistency, yao2023tree}. As a result, recent work has explored more efficient reasoning paradigms that reduce dependence on explicit textual rationales~\citep{he2026latent, li2025implicit}. 

Implicit chain-of-thought (iCoT) encodes reasoning in latent representations rather than explicit text~\citep{deng2024internal, hao2024cocnut}. Recent iCoT distillation approaches transfer explicit rationales into latent reasoning states~\citep{codi2025,pccot2025,kava2025,simcot2026}, but still underperform explicit CoT on difficult mathematical reasoning tasks. A key challenge is that explicit reasoning steps lack a clear correspondence to latent dynamics, making token-to-latent transfer inherently ill-posed. Existing methods rely on local supervision~\citep{codi2025} or sampled intermediate states~\citep{kava2025}, which may not fully capture the global reasoning trajectory.

\begin{figure*}[t]
    \centering

    \begin{subfigure}[t]{0.4\textwidth}
        \centering
        \includegraphics[width=\linewidth]{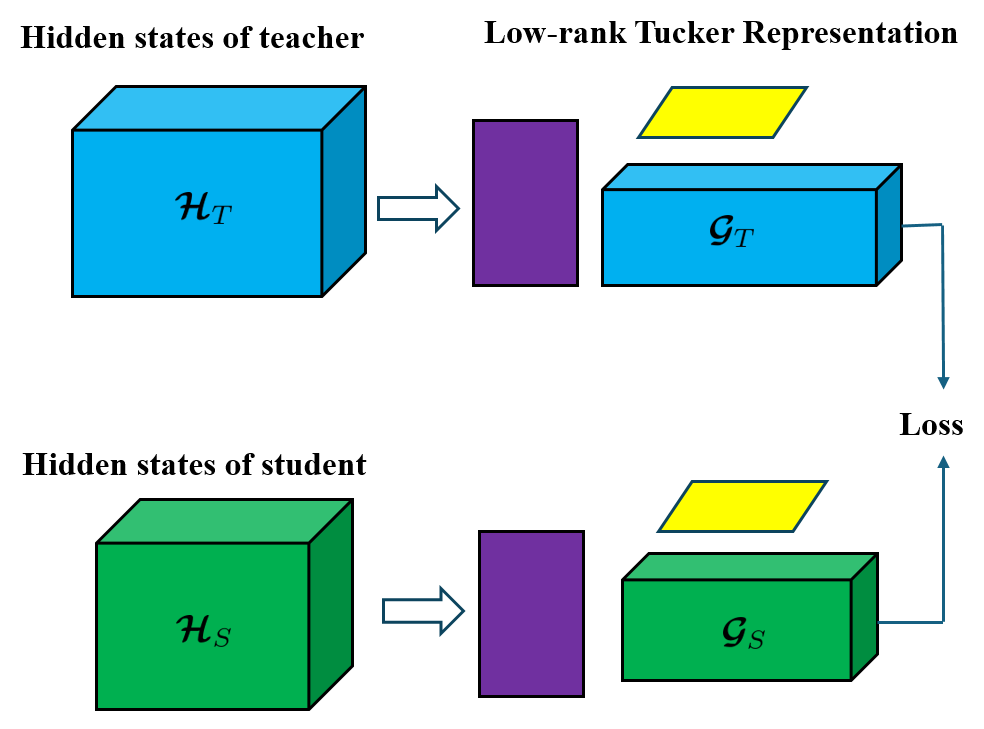}
        \caption{High-level overview of LoRi.}
        \label{fig:lori_framework}
    \end{subfigure}
    \hfill
    \begin{subfigure}[t]{0.55\textwidth}
        \centering
        \includegraphics[width=\linewidth]{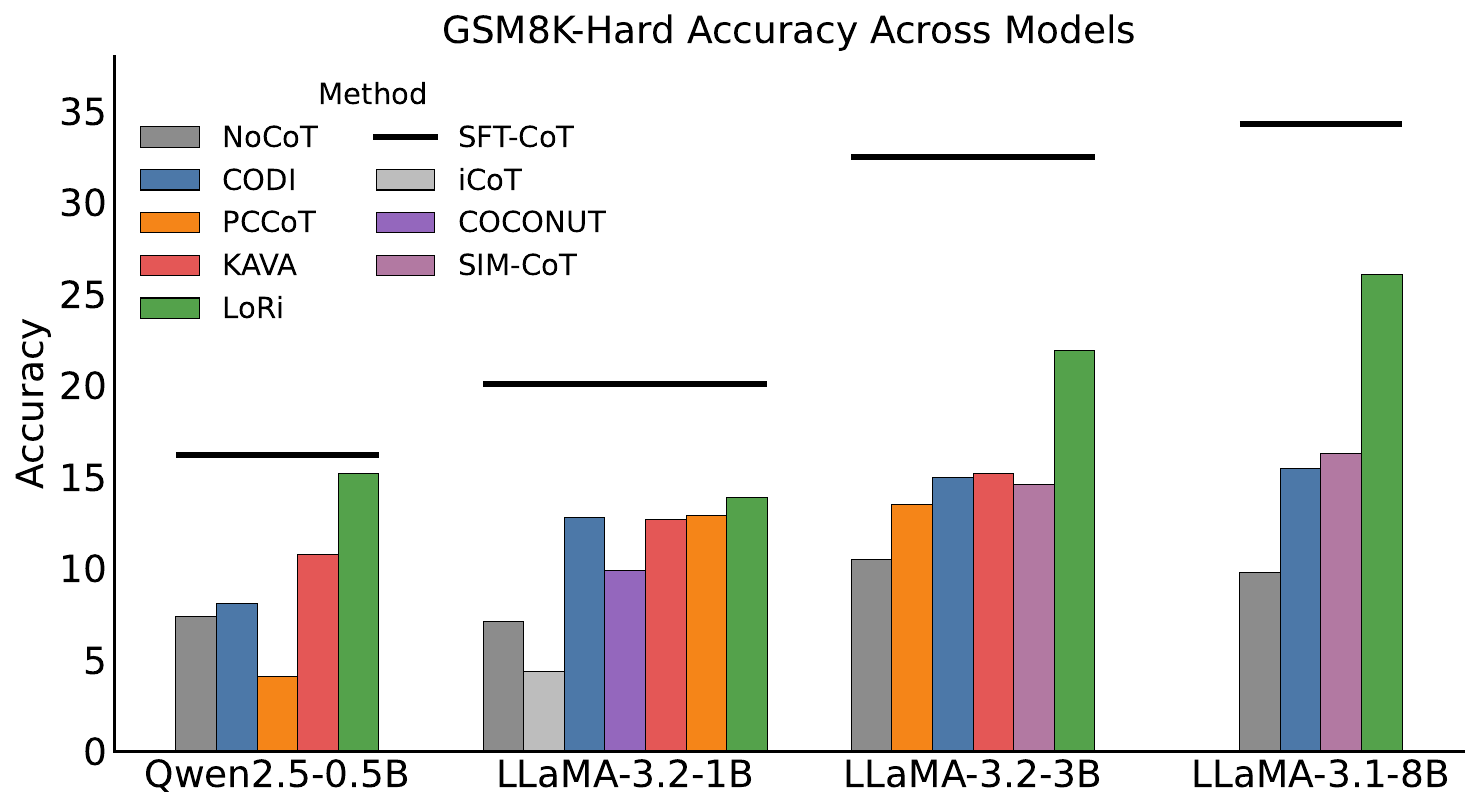}
        \caption{Representative reasoning results on GSM8K-Hard.}
        \label{fig:gsm8k_hard_results}
    \end{subfigure}

    \caption{
    Overview of the proposed low-rank iCoT distillation framework and its reasoning performance. (a) Low-rank factors learned from the teacher hidden states are used to project teacher and student representations into a shared low-rank subspace. The projected low-rank representations are then matched through a loss function. (b) LoRi improves reasoning accuracy over prior implicit CoT baselines across model scales on GSM8K-Hard. }
    \label{fig:summary}
\end{figure*}

Prior work suggests that model representations exhibit low-dimensional geometric structure~\citep{yu2023lowrank,golowich2025lowrank,modell2025manifold,park2024linear}. While recent studies analyze reasoning through trajectory geometry and separability~\citep{sun2026subspace,zhou2026geometry,li2025rema}, they do not directly examine the low-rank structure of hidden states across tokens and layers. By stacking hidden states across layers and CoT tokens, we find that the normalized cumulative singular values grow rapidly with rank (Appendix~\ref{app:motivation}), indicating that reasoning trajectories are well-approximated by low-dimensional subspaces.

Motivated by this observation, we propose a low-rank iCoT distillation framework that aligns the global geometry of the teacher’s reasoning trajectory through low-rank statistical representations. Instead of mimicking explicit reasoning steps, the student learns the low-dimensional subspace underlying the teacher’s reasoning dynamics [Fig.~\ref{fig:summary} (a)]. This allows the student to capture the principal reasoning structure with a short latent trajectory, enabling length-invariant and efficient distillation.

Our main contributions are summarized below:

\begin{itemize}[leftmargin=*]

\item \textbf{Low-rank iCoT distillation.}
We propose an iCoT distillation framework that transfers reasoning by aligning teacher and student trajectories in a shared low-rank subspace using first- and second-order hidden-state statistics.

\item \textbf{Efficient, length-invariant global reasoning transfer.}
The formulation transfers long CoT reasoning into short latent trajectories independent of sequence length, yielding efficient reasoning with a small number of latent steps and without requiring additional intermediate supervision.

 \item \textbf{Consistent gains across models and benchmarks.}
Across models and scales, LoRi consistently outperforms prior iCoT methods, improving accuracy by up to $\sim$12\% and achieving strong gains on GSM8K-Hard [up to $\sim$10\%, as shown in Fig.~\ref{fig:summary} (b)]. At larger scales, LoRi substantially narrows the gap to explicit CoT.

\end{itemize}

\section{Background and Related Work}

\paragraph{iCOT Distillation.}
In reasoning, a model predicts the final answer $\vect{y}$ conditioned on an intermediate reasoning process $\vect{\tau}$ given an input  $\vect{x}$,
\[
p(\vect{y},\vect{\tau} \mid \vect{x})=p(\vect{\tau} \mid \vect{x})\,p(\vect{y} \mid \vect{x},\vect{\tau}),
\]
where $\vect{\tau}$ may correspond to explicit textual rationales (explicit CoT) or implicit internal reasoning processes (iCoT). Following prior work~\citep{codi2025,kava2025,simcot2026}, we consider a teacher--student distillation framework to transfer reasoning from CoT into iCoT.
\begin{itemize}[leftmargin=*]
    \item {\bf Teacher Model.} The teacher generates a natural language reasoning trajectory
$\vect{r} = (r_1,\cdots,r_N)$ and defines the joint conditional distribution
\[
p_T(\vect{y},\vect{r} \mid \vect{x}) = p_T(\vect{r} \mid \vect{x})\, p_T(\vect{y} \mid \vect{x},\vect{r}).
\]

\item {\bf Student Model.} The student constructs a latent reasoning trajectory $\{\vect{z}_t\}_{t=1}^{K}$, where $K \ll N$ and each $\vect{z}_t$ represents a hidden reasoning state. The trajectory is generated recursively as
\[
\vect{z}_t = f_{\vect{\theta}}(\vect{x}, \vect{z}_{1:t-1})\,, \qquad t = 1,\ldots,K,
\]
where $f_{\vect{\theta}}$ is the student model, with $\vect{z}_1$ initialized from $\vect{x}$. The final answer is then generated autoregressively:
\[
p_S(\vect{y} \mid \vect{x}, \vect{z}_1,\ldots,\vect{z}_K).
\]

\item{\bf Training.} The student is trained through an objective of the form
\[
\mathcal{L}
=
\mathcal{L}_{\mathrm{reason}}
+
\lambda \mathcal{L}_{\mathrm{task}},
\]
where $\mathcal{L}_{\mathrm{reason}}$ transfers reasoning behavior from explicit CoT teacher into latent reasoning dynamics, $\mathcal{L}_{\mathrm{task}}$ supervises answer prediction, and $\lambda$ balances the two objectives.

\end{itemize}
Although iCoT improves inference efficiency, it often reduces reasoning accuracy. Existing methods mainly differ in how they transfer reasoning from explicit CoT into iCoT. Stepwise Internalization~\citep{deng2024internal} gradually removes CoT tokens through iterative fine-tuning, while COCONUT~\citep{hao2024cocnut} replaces textual reasoning with hidden-state dynamics. Recent approaches rely on distillation~\citep{deng2024implicit}: CODI~\citep{codi2025} distills reasoning at the answer boundary, PCCoT~\citep{pccot2025} adds tokens for intermediate reasoning states, KAVA~\citep{kava2025} aligns KV-cache dynamics, and SIM-CoT~\citep{simcot2026} uses step-level supervision to guide reasoning trajectories.

\paragraph{Low-Dimensional Structure.}
The manifold hypothesis states that high-dimensional data often lie on a low-dimensional manifold capturing their intrinsic structure~\citep{bengio2013representation, fefferman2016testing,chen2022self}. Recent work suggests that LLM representations exhibit similar low-dimensional geometry, including approximately low-rank structure of activations~\citep{yu2023lowrank,chen2024pid,liu2025cola}, linear semantic directions~\citep{park2024linear}, representation manifold structure~\citep{modell2025manifold}, and low-rank behavior in output logits~\citep{golowich2025lowrank}. 

\paragraph{Reasoning Geometry.} In reasoning tasks,~\citet{sun2026subspace} show that reasoning trajectories pass through step-specific subspaces that become increasingly separable with depth. \citet{zhou2026geometry} characterize reasoning as smooth flows shaped by logical structure, while \citet{li2025rema} propose a reasoning manifold where correct trajectories concentrate in low-dimensional regions and errors arise from deviations. Together, these studies suggest that high-dimensional reasoning trajectories can be effectively approximated by low-dimensional subspaces or manifolds.

\section{The \textbf{LoRi} Method}

We view the teacher's reasoning process as a trajectory in hidden-state space that lies near a low-dimensional subspace. From this perspective, distillation need not enforce step-by-step correspondence; instead, it should align the student's latent reasoning dynamics with the dominant structure of the teacher's trajectory. Motivated by this view, we propose \textbf{LoRi} (Low-Rank iCoT), a distillation framework that transfers reasoning from explicit chain-of-thought into a compact implicit latent process. LoRi combines two complementary objectives: rationale-level alignment, which preserves the global geometry of the teacher's trajectory, and anchor-level alignment, which regularizes the transition from latent reasoning to answer generation. Together with an efficient training scheme based on precomputed low-rank representations, LoRi enables scalable, length-invariant distillation from a CoT teacher to an iCoT student.

\begin{figure*}[t]
    \centering
    \includegraphics[width=\textwidth]{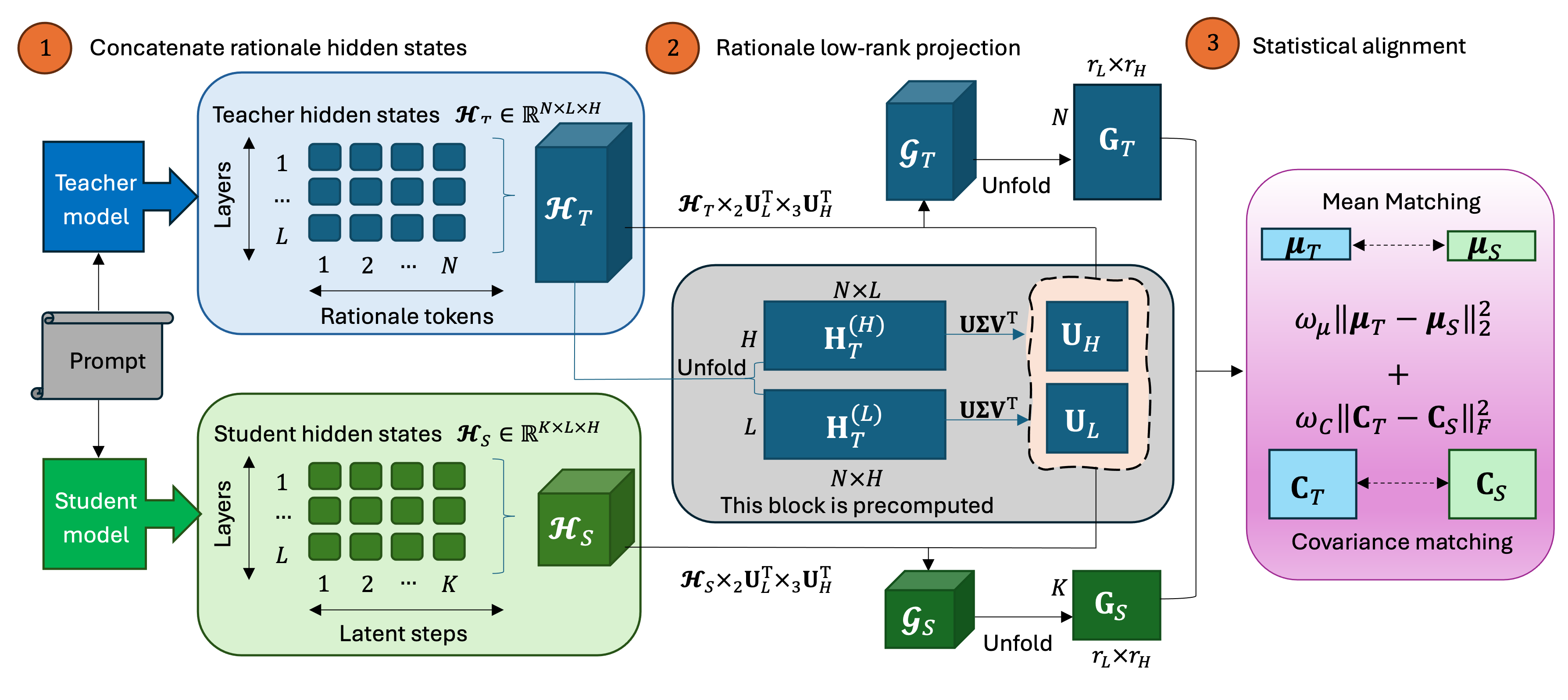}
    \caption{
    Overview of low-rank rationale-level alignment in LoRi.
    }
    \label{fig:rationale_fig}
\end{figure*}

\subsection{Low-Rank iCoT Distillation}
\label{sec:lr-distil}
\paragraph{Distillation loss.} The goal of distillation is to transfer the reasoning structure encoded in the teacher's explicit rationales $r$ into the student's latent reasoning dynamics $z$, while preserving the student's ability to generate explicit step-by-step solutions and final answers. We train the student with the composite objective
\[
\mathcal{L} = \mathcal{L}_{\mathrm{LR}}+\lambda\,\mathcal{L}_{\mathrm{CE}},
\label{eq:loss}
\]
where $\mathcal{L}_{\mathrm{CE}}$ is a cross entropy that supervises the student's explicit output sequence, $\mathcal{L}_{\mathrm{LR}}$ aligns the student's latent reasoning states with the teacher's hidden representations in a low-rank subspace, and $\lambda$ balances the two terms. Intuitively, $\mathcal{L}_{\mathrm{LR}}$ encourages both models to share the dominant low-rank structure of the teacher's reasoning trajectory, improving the student's stability and generalization.

\paragraph{Low-rank term.} We define $\mathcal{L}_{\mathrm{LR}}$ as the sum of two complementary components,
\[
\mathcal{L}_{\mathrm{LR}}
=
\mathcal{L}_{\mathrm{rationale}}
+
\beta \,\mathcal{L}_{\mathrm{anchor}},
\]
where $\mathcal{L}_{\mathrm{rationale}}$ aligns the student's latent reasoning dynamics with the teacher's hidden states over the full reasoning trajectory, and $\mathcal{L}_{\mathrm{anchor}}$ aligns the student's representation at the answer prediction position with the
teacher's corresponding hidden state, while $\beta$ controlling its relative weight. The former captures the global low-rank structure of the teacher's reasoning process, while the latter provides a localized signal that guides the transition from internal reasoning to answer generation.

\subsection{Rationale Level Alignment ($\mathcal{L}_{\mathrm{rationale}}$)}
\label{sec:rational-align}

To transfer reasoning capability, we align the student's hidden states with the teacher's reasoning trajectory in a low-rank subspace. Let $\ten{H}_T \in \mathbb{R}^{N \times L \times H}$ denote the teacher's hidden states for rationale tokens, where $L$ is the number of layers and $H$ is the hidden dimension. Similarly, let $\ten{H}_S \in \mathbb{R}^{K \times L \times H}$ be the student's hidden states for its implicit reasoning steps. Instead of directly matching these high-dimensional tensors, we project them into a shared low-rank subspace and align their low-dimensional representations through a statistical matching objective (Fig.~\ref{fig:rationale_fig}).

\paragraph{Tucker Representation for Reasoning Trajectories.} 
According to the Tucker decomposition~\citep{delathauwer2000hosvd}, a tensor $\ten{X} \in \mathbb{R}^{N\times L \times H}$ is factorized into a tensor core $\ten{Q} \in \mathbb{R}^{r_N \times r_L \times r_H}$ and orthonormal factor matrices $\mat{U}_N \in \mathbb{R}^{N\times r_N}$, $\mat{U}_L \in \mathbb{R}^{L\times r_L}$, $\mat{U}_H \in \mathbb{R}^{H\times r_H}$:
\[
\ten{X} \approx \ten{Q} \times_{1} \mat{U}_N \times_{2} \mat{U}_L \times_{3} \mat{U}_H,
\]
where $\times_n$ denotes the mode-$n$ tensor product (see Appendix~\ref{app:tensor_contraction}). We construct a low-rank Tucker-style representation of the teacher's
hidden states along the layer and hidden dimensions by
extracting low-rank factor matrices via SVD (top left singular vectors) of unfolded $\ten{H}_T$ along
the layer and hidden dimensions, yielding orthonormal column matrices
$\mat{U}_L \in \mathbb{R}^{L \times r_L}$ and 
$\mat{U}_H \in \mathbb{R}^{H \times r_H}$, respectively, which
capture the dominant subspaces of the teacher's hidden-state representations. Each reasoning state is then projected as
\[
\ten{G} = \ten{X} \times_2 \mat{U}_L^\top \times_3 \mat{U}_H^\top,
\]
where $\ten{X}$ denotes either $\ten{H}_T$ or $\ten{H}_S$.
 This yields low-dimensional representations
\(\ten{G}_T \in \mathbb{R}^{N \times r_L \times r_H}\) for the teacher
and \(\ten{G}_S \in \mathbb{R}^{K \times r_L \times r_H}\) for the
student. We then reshape these tensors along the last two modes to obtain
matrix representations
\[
\mat{G}_T \in \mathbb{R}^{N \times (r_L r_H)},
\qquad
\mat{G}_S \in \mathbb{R}^{K \times (r_L r_H)},
\]
where each row corresponds to a reasoning step represented in the
shared low-rank subspace.

\begin{figure*}[t]
    \centering
    \includegraphics[width=\textwidth]{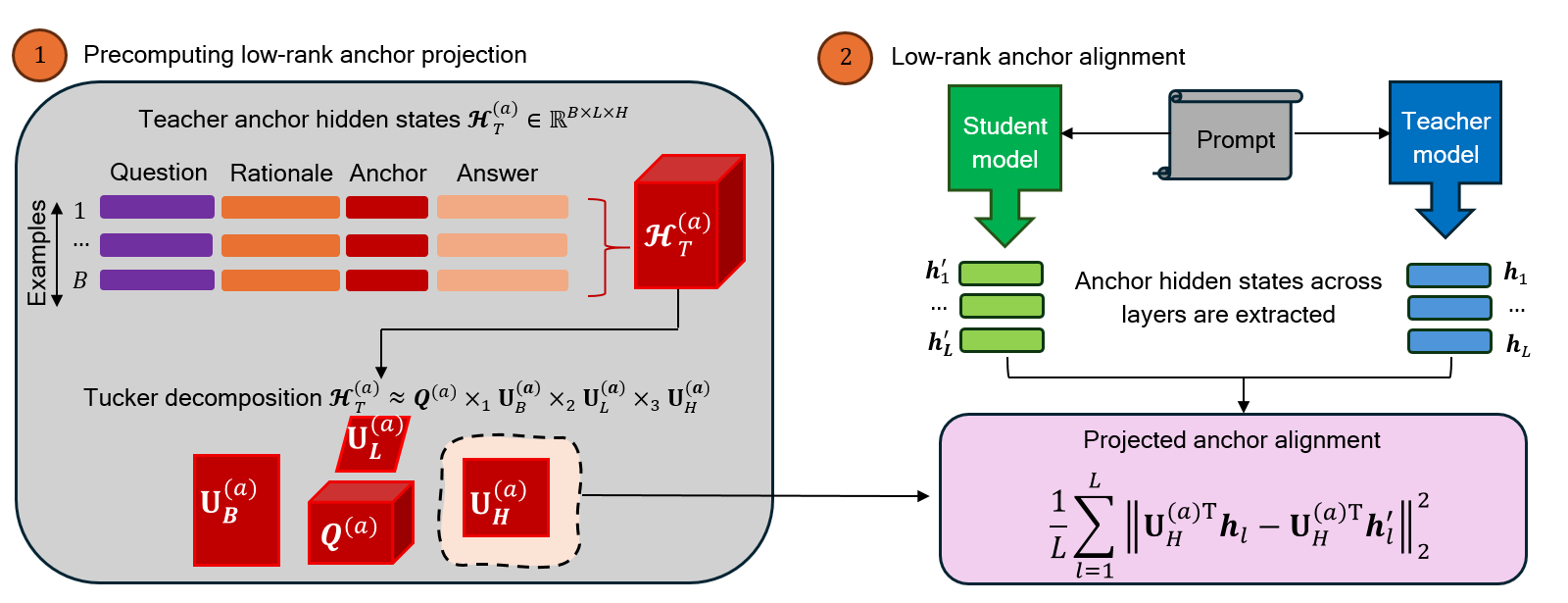}
    \caption{
    Overview of low-rank anchor alignment in LoRi.
    }
    \label{fig:anchor_fig}
\end{figure*}

\paragraph{Statistical Alignment in Low-rank Space.} We align student and teacher representations by matching the first- and second-order statistics of their projections in a shared low-rank subspace. For
\(\mat{G} \in \mathbb{R}^{M \times (r_L r_H)}\), where \(M=N\) for the
teacher and \(M=K\) for the student, we define
\[
\boldsymbol{\mu} = \frac{1}{M}\sum_{m=1}^{M}\mat{G}_{m,:}, \qquad
\mat{C} = \frac{1}{M}\mat{G}^\top \mat{G}.
\]
Let
\((\mat{C}_T,\boldsymbol{\mu}_T)\) and \((\mat{C}_S,\boldsymbol{\mu}_S)\) denote the statistics computed from \(\mat{G}_T\) and
\(\mat{G}_S\), respectively. We define the rationale-level loss as
\[
\mathcal{L}_{\mathrm{rationale}}
=
\omega_C\,\left\| \mat{C}_S - \mat{C}_T \right\|_F^2
\,+
\,\omega_{\mu}\,\left\| \boldsymbol{\mu}_S - \boldsymbol{\mu}_T \right\|_2^2,
\]
where $\omega_C$ and $\omega_{\mu}$ weight covariance and mean alignment. Matching covariance aligns the principal low-rank structure, while matching means aligns the trajectory centroids. Together, these constraints guide the student toward the teacher's reasoning geometry and discourage degenerate solutions such as collapsed latent trajectories.

This objective transfers the global structure of the teacher's reasoning process without token-level alignment and remains invariant to reasoning length. The low-rank factors $\mat{U}_L$ and $\mat{U}_H$ are precomputed from the teacher and fixed during training, enabling efficient projection  without repeated factorization.

\subsection{Anchor-Level Alignment ($\mathcal{L}_{\mathrm{anchor}}$)}
\label{sec:anchor-align}

While the rationale-level objective captures global reasoning structure, it does not explicitly constrain the transition from latent reasoning to answer generation. To address this, we introduce a localized alignment term at the answer prediction position. This position corresponds to a fixed prompt-aligned token (e.g., ``The final answer is''), which serves as a consistent transition between reasoning and answer generation rather than part of the reasoning trajectory itself (Figure~\ref{fig:anchor_fig}).

Let $\ten{H}_T^{\mathrm{anchor}} \in \mathbb{R}^{B \times L \times H}$
denote the collection of teacher hidden states at the answer prediction
position across $B$ training samples. We construct a low-rank Tucker
decomposition of this tensor, yielding factor matrices
\[
\mat{U}_B^{(a)} \in \mathbb{R}^{B \times r_B}, \,
\mat{U}_L^{(a)} \in \mathbb{R}^{L \times r_L}, \,
\mat{U}_H^{(a)} \in \mathbb{R}^{H \times r_H}.
\]

In our formulation, we use the hidden-mode factor $\mat{U}_H^{(a)}$, which captures the dominant subspace of the teacher's hidden representations at the anchor position across the dataset. For a given sample, let $\vect{h}_\ell, \vect{h}_\ell' \in \mathbb{R}^{H}$,
for $\ell = 1,\ldots,L$, denote the teacher and student hidden states at
the answer token across layers. The anchor-level loss is defined as
\[
\mathcal{L}_{\mathrm{anchor}} =
\frac{1}{L}\sum_{\ell=1}^{L}
\left\|
\mat{U}_H^{(a)\top} \vect{h}_\ell
-
\mat{U}_H^{(a)\top} \vect{h}_\ell'
\right\|_2^2.
\]
This formulation leverages a shared low-rank subspace learned across
training samples, ensuring that the student's representation at the
answer prediction point aligns with the dominant structure of the
teacher's hidden states at that location, while remaining complementary
to the global alignment enforced by
$\mathcal{L}_{\mathrm{rationale}}$.

\subsection{Implications for Reasoning}

\paragraph{Reasoning compression.}
The low-rank structure provides a geometric explanation for why long
chain-of-thought reasoning can be compressed into a shorter latent
trajectory. Since the teacher's hidden-state trajectory lies near a
low-rank subspace, its dominant variation can be represented using
a limited number of degrees of freedom, largely independent of sequence length. By aligning the student within this subspace, the student is encouraged to reproduce the principal reasoning dynamics without explicitly matching each reasoning step.

\paragraph{Length invariance.}
Importantly, our formulation is invariant to the length of the reasoning trajectory. Since alignment is performed through aggregated statistics rather than step-wise correspondence, the student can construct a shorter
latent trajectory that matches the global geometry of the teacher. This provides an intuitive justification for transferring reasoning from long explicit CoT sequences to compact implicit reasoning processes.

\begin{table*}[t]
\small
\centering
\caption{Accuracy (\%) of different reasoning methods on math benchmarks across models. The best iCoT method is shown in bold, and the second-best method is underlined. Explicit CoT results are shown in gray. Parentheses denote CODI/KAVA results without their final-step dropping heuristic.}
\label{tab:reasoning_results}
\begin{tabular}{llcccc}
\toprule
\textbf{Model} & \textbf{Method} & \textbf{GSM8K} & \textbf{GSM8K-Hard} & \textbf{SVAMP} & \textbf{Average} \\
\midrule

\multirow{5}{*}{Qwen2.5-0.5B}
 & \textgray{SFT-CoT}     & \textgray{52.3} & \textgray{16.2} & \textgray{62.3} & \textgray{43.6} \\
 & NoCoT   & 31.5 & 7.4 & 34.5 & 24.5 \\
 & CODI    & 37.5 & 8.1 & 47.0 & 30.9 \\
 & PCCOT & 20.5 & 4.1 & 33.0 & 19.2 \\
 & KAVA & \underline{46.9} & \underline{10.8} & \underline{50.6} & \underline{36.1} \\
  \rowcolor{blue!05}
 & LoRi& \textbf{50.0} & \textbf{15.2} & \textbf{63.0} & \textbf{42.7} \\
 & \textgray{LoRi-E} & \textgray{52.0} & \textgray{16.9} & \textgray{62.7} & \textgray{43.9} \\
\midrule

\multirow{5}{*}{LLaMA-3.2-1B}
& \textgray{SFT-CoT}     & \textgray{58.4} & \textgray{20.1} & \textgray{61.0} & \textgray{46.5} \\
 & NoCoT   & 30.9& 7.1 & 44.1 & 27.4 \\
 & iCoT    & 19.0 & 4.4 & 40.9 & 21.4 \\
 & CODI    & 55.6 (47.2) & 12.8 & \underline{61.1} & \underline{43.2} \\
 & COCONUT & 45.3 & 9.9 & 48.8 & 34.7 \\
& KAVA & \textbf{56.5} (51.2) & 12.7 & 58.9 & 42.7 \\
& PCCoT & 54.2 & \underline{12.9} & 57.7 & 41.6\\
\rowcolor{blue!05}
& LoRi & \underline{54.9} & \textbf{13.9} & \textbf{64.0} & \textbf{44.3} \\
 & \textgray{LoRi-E} & \textgray{58.1} & \textgray{14.9} & \textgray{64.3} & \textgray{45.8} \\
\midrule

\multirow{5}{*}{LLaMA-3.2-3B}
 & \textgray{SFT-CoT}     & \textgray{74.9} & \textgray{32.5} & \textgray{78.7} & \textgray{62.0} \\
 & NoCoT   & 41.7 & 10.5 & 56.9 & 36.4 \\
 & PCCoT    & 54.7 & 13.5 & 69.5 & 45.9 \\
 & CODI    & 61.0 & 15.0 & 72.4 & 49.5 \\
 & KAVA & \underline{65.7} & \underline{15.2} & 72.7 & \underline{51.2} \\
& SIM-CoT& 62.3 & 14.6 & \underline{74.9} & 50.6 \\
\rowcolor{blue!05}
& LoRi & \textbf{70.0} & \textbf{21.9} & \textbf{75.0} & \textbf{55.6} \\
 & \textgray{LoRi-E} & \textgray{74.3} & \textgray{24.1} & \textgray{79.3} & \textgray{59.2} \\
\midrule

\multirow{5}{*}{LLaMA-3.1-8B}
& \textgray{SFT-CoT}     & \textgray{76.6} & \textgray{34.3} & \textgray{81.0} & \textgray{64.0} \\
 & NoCoT   & 39.5 & 9.8 & 55.3 & 34.9 \\
 & CODI    & 61.1 & 15.5 & 78.1 & 51.6 \\
 & SIM-CoT & \underline{64.1} & \underline{16.3} & \underline{79.4} & \underline{53.3} \\
 \rowcolor{blue!05}
 &  LoRi & \textbf{79.3} & \textbf{26.1} & \textbf{83.3} & \textbf{62.9} \\
 & \textgray{LoRi-E} & \textgray{80.4} & \textgray{27.2} & \textgray{83.3} & \textgray{63.6} \\
\bottomrule
\end{tabular}
\end{table*}

\subsection{Relation to Prior Work}

Our method differs from prior iCoT distillation approaches in both supervision and training. In the following we compare LoRi with representative prior work CODI~\citep{codi2025}, KAVA~\citep{kava2025} and SIM-CoT~\citep{simcot2026}:

\begin{itemize}[leftmargin=*]
    \item CODI aligns representations at a single boundary token, reducing supervision to a point-wise constraint and overlooking the structure of the reasoning trajectory. As a result, it struggles to transfer multi-step reasoning dynamics. In contrast, our method captures the full reasoning trajectory through a low-rank representation, supplemented by a localized anchor term that constrains the transition from reasoning to answer generation at a fixed answer-boundary position.
    
    \item KAVA distills teacher KV-cache trajectories through step-wise alignment in KV space using sampled rationale tokens. By contrast, LoRi models the reasoning process through low-rank hidden-state geometry, enabling length-invariant distillation without token-level sampling.

    \item SIM-CoT~\citep{simcot2026} applies step-level supervision by aligning latent states with explicit reasoning tokens via an auxiliary decoder. Our method instead avoids explicit step-wise alignment and supervises the student through low-rank trajectory structure.
\end{itemize}

Finally, existing iCoT distillation methods rely on joint teacher--student training with online teacher inference. In contrast, we adopt a two-stage procedure: teacher-derived low-rank factors are precomputed once, after which the student is fine-tuned independently on a subset of the training data. This substantially reduces training cost while preserving effective reasoning transfer.

\section{Results}
\label{sec: result}

\subsection{Experimental Setup}


Following prior work, we evaluate LoRi on standard mathematical reasoning benchmarks, which provide rigorous tests of multi-step reasoning through structured problem solving and verifiable answers. This focus is also motivated by prior findings that CoT is most effective for mathematical and symbolic reasoning~\citep{sprague2024cot_or_not}. Our additional analysis across math, coding, science, and commonsense tasks further supports this observation with CoT-SFT providing little or no advantage over standard SFT outside mathematical reasoning (Appendix~\ref{app:extend_beyond_math}).

We compare LoRi against NoCoT, CODI~\citep{codi2025}, KAVA~\citep{kava2025}, SIM-CoT~\citep{simcot2026}, PCCoT~\citep{pccot2025}, COCONUT~\citep{hao2024cocnut}, and SFT-CoT across Qwen~\citep{qwen2023} and LLaMA~\citep{llama3_2024} model families. To study scalability, we evaluate models ranging from 0.5B to 8B parameters. SFT-CoT serves as an upper-bound reference since it relies on explicit reasoning at inference time, whereas LoRi targets implicit reasoning.

The teacher model is first fine-tuned on the full GSM8K-Aug dataset~\citep{deng2024internal,deng2024implicit} and then frozen during distillation. The student is initialized from the same base model. A random 128-sample subset of GSM8K-Aug is used to estimate the teacher-derived low-rank quantities, which are subsequently kept fixed during student training. We report final-answer accuracy based on extracted numerical outputs. Additional implementation details and hyperparameters are provided in Appendix~\ref{app:hyperparams}.

\subsection{Accuracy Comparison}

Table~\ref{tab:reasoning_results} summarizes performance on several benchmarks and model scales. Overall, LoRi consistently outperforms prior implicit reasoning methods and substantially narrows the gap to full CoT models. This suggests that low-rank alignment captures global reasoning dependencies that token-level methods fail to model.

\paragraph{Performance w.r.t. Model Scale.}
For small models, LoRi provides strong gains over prior implicit reasoning methods. 
On Qwen2.5-0.5B, LoRi achieves an average accuracy of $42.7\%$, substantially outperforming CODI ($30.9\%$), KAVA ($36.1\%$), and PCCoT ($19.2\%$). On LLaMA-3.2-1B, LoRi reaches the best average accuracy among iCoT methods ($44.3\%$), although its advantage over CODI and KAVA is more nuanced: CODI and KAVA obtain slightly higher accuracy on GSM8K, while LoRi performs better on GSM8K-Hard and SVAMP. Moreover, when CODI and KAVA are evaluated without their final-step dropping heuristic~\citep{kava2025}, LoRi outperforms them by a clear margin. 
This is notable because LoRi does not use step-level supervision or heuristic rationale truncation.

For larger models, the gains are pronounced. On LLaMA-3.2-3B, LoRi achieves $55.6\%$ average accuracy, outperforming all implicit baselines by a clear margin. On LLaMA-3.1-8B, LoRi reaches $62.9\%$, approaching the performance of full CoT models ($64.0\%$). These results suggest that the proposed method scales effectively, substantially narrowing the gap between implicit and explicit reasoning.

\paragraph{Hard Reasoning Tasks.} The improvements are particularly notable on GSM8K-Hard, which requires more complex multi-step reasoning. For example, on LLaMA-3.2-3B, LoRi improves performance from $15.2\%$ (KAVA) to $21.9\%$; on LLaMA-3.1-8B, LoRi achieves $26.1\%$, substantially outperforming CODI ($15.5\%$) and SIM-CoT ($16.3\%$). This suggests that capturing the global structure of reasoning through low-rank alignment is especially beneficial for harder reasoning tasks.

\paragraph{Capability of Explicit Reasoning.} To evaluate whether LoRi preserves explicit reasoning ability, we test LoRi-E, which uses the same student model but performs standard CoT generation at inference time. LoRi-E achieves performance comparable to SFT-CoT across all models and benchmarks. For instance, on LLaMA-3.1-8B, LoRi-E attains an average accuracy of $63.6\%$, close to $64.0\%$ for SFT-CoT. These results suggest that the proposed distillation method preserves explicit reasoning capability while enabling both implicit and explicit reasoning at inference time.

\subsection{Training Efficiency}
\begin{table}[t]
\centering
\small
\caption{Training cost on NVIDIA RTX 6000 GPUs. FT: teacher fine-tuning. Distill: reasoning distillation.}
\label{tab:training_cost}
\setlength{\tabcolsep}{4pt}
\begin{tabular}{lccc}
\toprule
Model & Method & Stage & GPU-h \\
\midrule
\multirow{2}{*}{LLaMA-1B}
& CODI & Distill & 16.5 \\
& LoRi & FT + Distill & \textbf{4.75 + 0.02} \\
\midrule
\multirow{2}{*}{LLaMA-3B}
& CODI & Distill & 30.9 \\
& LoRi & FT + Distill & \textbf{8.8 + 0.03} \\
\bottomrule
\end{tabular}
\end{table}

Beyond accuracy, LoRi also 
improves training efficiency over prior implicit reasoning methods. Existing iCoT distillation approaches, including CODI, KAVA, and SIM-CoT, rely on joint teacher--student training with repeated teacher forward passes to obtain intermediate representations, resulting in substantial computational overhead. In contrast, LoRi uses a two-stage procedure: teacher-derived low-rank factors are estimated once from a small subset of the training data and subsequently kept fixed during student optimization, avoiding repeated teacher forward passes. Despite this simplified pipeline, LoRi maintains strong benchmark performance while significantly improving training efficiency and scalability as summarized in Table~\ref{tab:training_cost}. In practice, the distillation overhead becomes effectively negligible, making LoRi a more scalable and accessible approach for implicit reasoning distillation. We use CODI as a representative baseline for training-cost reference since related iCoT distillation methods (e.g., PCCoT, KAVA, and SIM-CoT) employ similar joint teacher--student training procedures with repeated teacher forward passes, leading to very similar computational overhead.

\begin{figure}[t]
    \centering
    \includegraphics[width=\columnwidth]{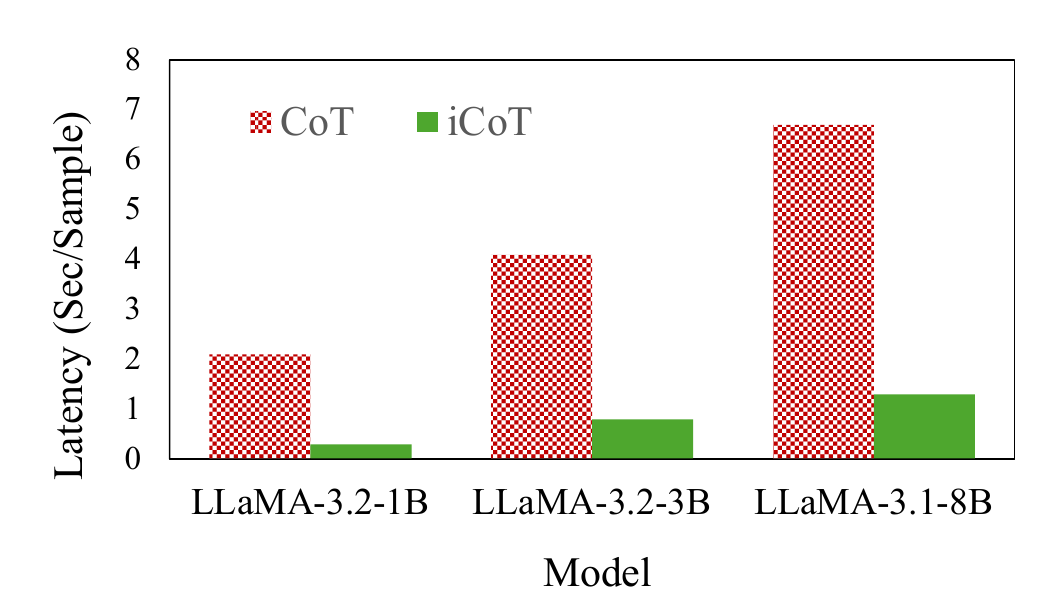}
    \caption{Inference latency comparison. Note: LoRi and prior iCoT methods have almost the same inference cost since they use the same latent-step inference procedure.}
    \label{fig:latency}
    \vspace{-10pt}
\end{figure}

\begin{figure*}[t]
    \centering
    \begin{subfigure}[t]{0.45\linewidth}
        \centering
        \includegraphics[width=\linewidth]{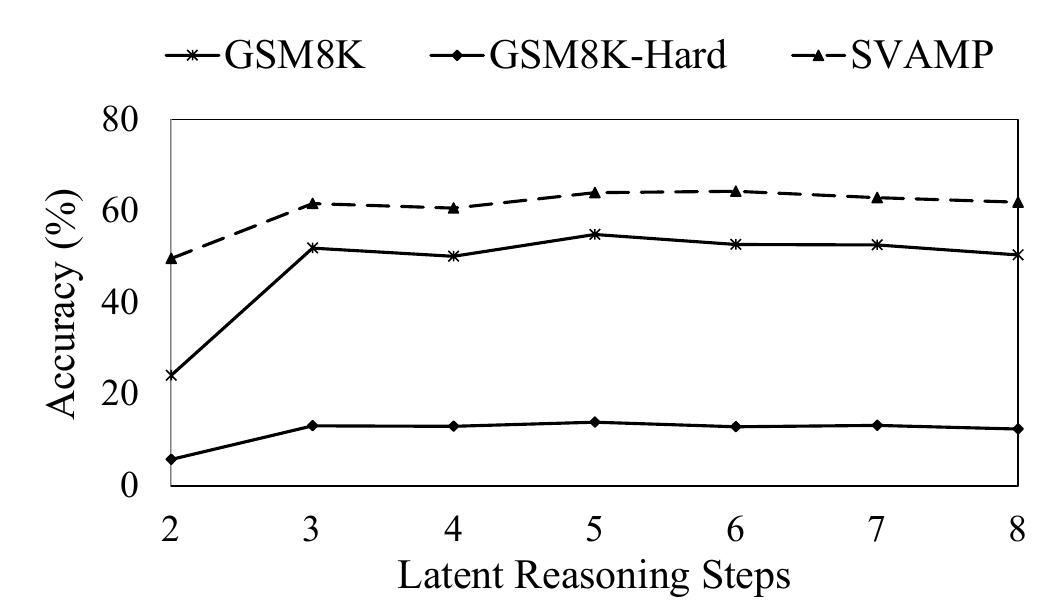}
        \label{fig:abl_num_steps}
    \end{subfigure}
    \hfill
    \begin{subfigure}[t]{0.45\linewidth}
        \centering
        \includegraphics[width=\linewidth]{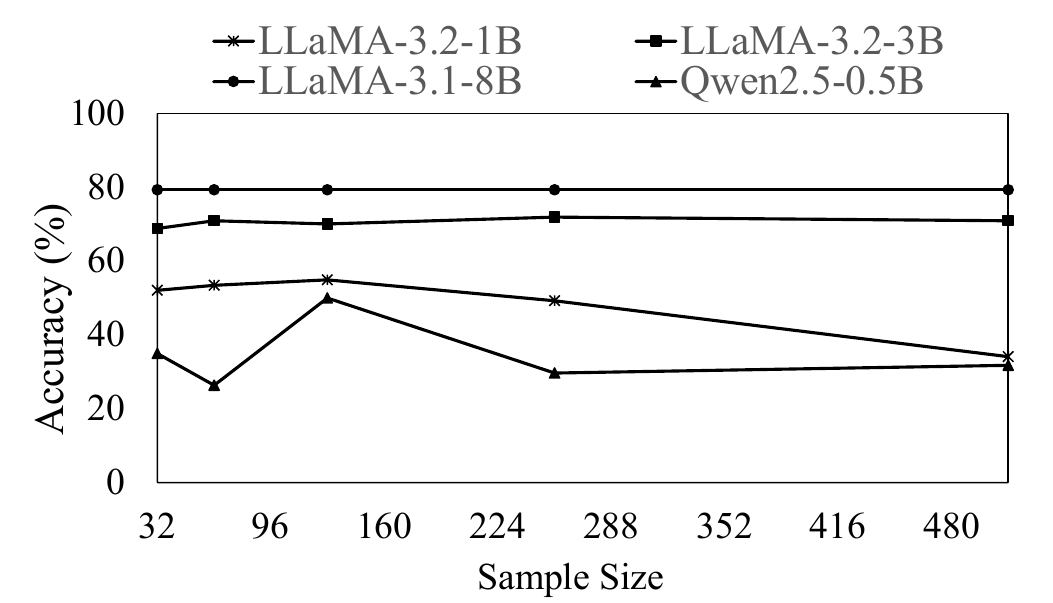}
        \label{fig:abl_sample_size}
    \end{subfigure}
    \caption{Ablation studies on reasoning steps (left) and training sample size (right).}
    \label{fig:abl}
    \vspace{-10pt}
\end{figure*}

\subsection{Inference Latency}

We compare the inference latency of LoRi-based iCoT and explicit CoT. Latency is measured as average wall-clock time per sample. CoT generates full reasoning sequences autoregressively, whereas iCoT uses a fixed number of latent steps ($K=5$) followed by short answer generation, significantly reducing decoding cost. As shown in Figure~\ref{fig:latency}, iCoT consistently achieves lower latency across all model scales. Specifically, it is about 6.9$\times$ faster for LLaMA-3.2-1B and 5.1$\times$ faster for both LLaMA-3.2-3B and LLaMA-3.1-8B. These results confirm the efficiency advantages of iCoT while showing that LoRi preserves this benefit alongside improved reasoning performance. Since LoRi and prior iCoT methods use the same latent-step inference procedure, their inference complexity is almost the same; the main difference lies in the distillation strategy during training.

\subsection{Ablation Studies}

We study the effect of the number of latent reasoning steps $K$ and the number of training samples used for distillation on model performance.

\paragraph{Ablation on Latent Reasoning Steps.}
Figure~\ref{fig:abl} reports benchmark accuracy as the number of latent steps $K$ varies from 2 to 8. Performance consistently improves from $K=2$ to $K=5$, suggesting that a minimum number of latent reasoning iterations is needed to capture the underlying reasoning dynamics. The best results are achieved at $K=5$ across all datasets, indicating that this setting provides sufficient capacity to represent the dominant low-rank reasoning structure. Increasing $K$ beyond 5 yields no further gains and occasionally causes slight degradation, implying that additional steps introduce redundancy rather than useful computation. Overall, these results support the view that reasoning trajectories can be compressed into a small number of latent steps without sacrificing essential structure.

\paragraph{Ablation on Training Sample Size.}
Figure~\ref{fig:abl} shows that accuracy improves rapidly in the low-data regime and saturates after roughly 128 samples across all model scales, indicating that strong performance can be achieved with only a small subset of the training data. One explanation is that the effective degrees of freedom of the distillation problem are limited: although the teacher produces long chain-of-thought trajectories, the information needed for the student may lie in a low-dimensional subspace. Under this view, distillation only needs to recover the dominant reasoning structure rather than the full trajectory. Consequently, relatively few samples are sufficient to estimate this structure, leading to rapid gains followed by saturation. This observation supports our low-rank formulation and suggests that the reasoning signal relevant for distillation is highly compressible.

\section{Conclusion}

We have proposed LoRi, a low-rank formulation for iCoT distillation. Empirically, LoRi outperformed prior implicit reasoning methods across model families, scales, and benchmarks and reduced the gap of iCoT with explicit CoT while preserving the efficiency advantages of implicit reasoning. In addition, the proposed method significantly reduced computational overhead, making distillation lightweight and scalable. The results also support a geometric perspective of reasoning: long chain-of-thought trajectories can be effectively compressed by preserving their low-rank structure. This perspective provides a principled and efficient alternative to existing iCoT distillation approaches and suggests that the essential dynamics of reasoning are governed by a low-dimensional subspace that can be exploited for both learning and inference.

\section*{Limitations}
The proposed formulation is motivated by an empirical observation that hidden-state reasoning trajectories exhibit strong low-rank structures across models and benchmarks. While LoRi improves implicit reasoning, the theoretical relationship between low-rank representation geometry and reasoning capability is still not fully understood. Our empirical evaluation is primarily based on a single mathematical reasoning training setting; although the coding analysis supports the broader low-rank observation, the usefulness of CoT varies across tasks, and LoRi’s effectiveness may therefore depend on the dataset and teacher training setting. Future work could further investigate the geometry of reasoning trajectories and the role of low-dimensional structure in CoT reasoning from a theoretical perspective. Additionally, undisclosed pretraining data may influence the observed CoT and non-CoT behaviors.


\bibliography{main}

\clearpage

\appendix

\section{Low-rank Analysis of CoT Hidden States}
\label{app:motivation}
 \begin{figure}
    \centering
    \vspace{-10pt}
\includegraphics[width=\linewidth]{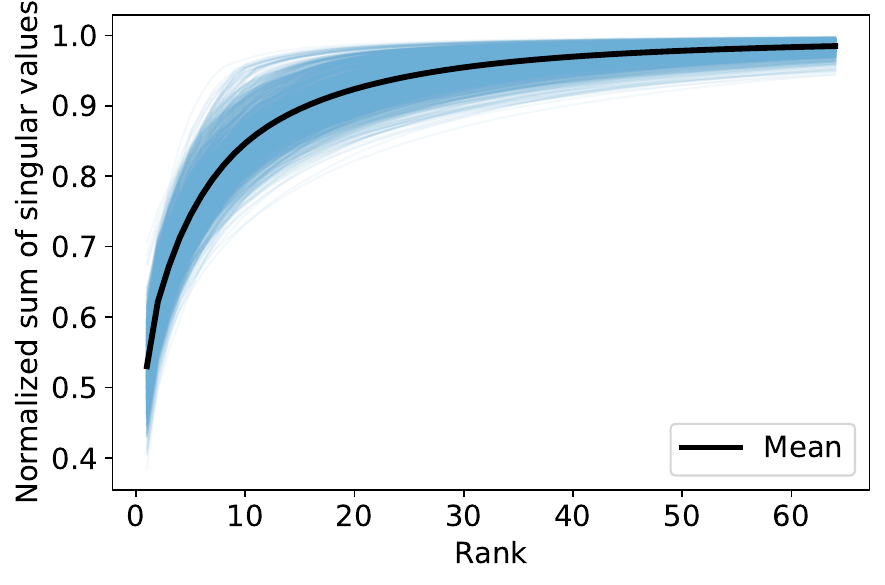}
    \caption{Normalized sum of singular values versus rank for stacked hidden states of all layers and all CoT tokens of LLaMA-3.2-1B model, where each curve corresponds to a single GSM8K example.}
    \label{fig:energy_rank}
\end{figure}

For this analysis, we stack the hidden states corresponding to all transformer layers and all CoT rationale tokens into a single matrix and compute its singular value decomposition (SVD). Figure~\ref{fig:energy_rank} shows the normalized cumulative singular values across GSM8K examples for LLaMA-3.2-1B. Despite the model having a hidden dimension of 2048, the singular values decay rapidly and are largely concentrated within a relatively small rank. This strong low-rank structure suggests significant compressibility in hidden-state reasoning dynamics, motivating the use of low-rank representations for implicit reasoning distillation.

We further analyze CoT hidden states for coding task using LLaMA-3.2-1B. As seen in Table~\ref{tab:coding_lowrank}, CoT fine-tuning on KodCode~\citep{kodcode} substantially concentrates the spectrum such that the dimensions required to capture 95\% and 99\% of the energy decrease from 293.3 to 85.3 and from 615.2 to 205.5, respectively, while the tail energy beyond rank 64 drops from 0.276 to 0.077. Meanwhile, accuracy improves on KodCode, MBPP~\citep{mbpp}, and HumanEval~\citep{humaneval}, suggesting that the increased spectral concentration reflects more structured representations rather than collapse. These results indicate that the low-rank structure may also emerge in non-mathematical reasoning.

\begin{table*}[t]
\small
\centering
\caption{Low-rank structure of coding reasoning trajectories using LLaMA-3.2-1B. 
We report the number of dimensions required to capture 95\% and 99\% of the spectral energy, the residual energy beyond rank 64, and accuracy (\%) on coding benchmarks.}
\label{tab:coding_lowrank}
\begin{tabular}{lcccccc}
\toprule
Variant
& Dims (95\%)
& Dims (99\%)
& Tail Energy ($r>64$)
& KodCode
& MBPP
& HumanEval \\
\midrule
Base CoT
& 293.3
& 615.2
& 0.276
& 15.5
& 35.5
& 21.3 \\
CoT-SFT (KodCode)
& 85.3
& 205.5
& 0.077
& 32.5
& 37.0
& 35.4 \\
\bottomrule
\end{tabular}
\end{table*}

\section{Ablation of the Anchor Loss Term}
\label{appx:loss_abl}

Table~\ref{tab:anchor_ablation} shows that the anchor-level loss $\mathcal{L}_{\mathrm{anchor}}$ generally improves performance, with the largest gains observed for smaller models. For Qwen2.5-0.5B, it yields substantial improvements on both GSM8K and SVAMP, indicating a strong additional supervision signal. For larger models, the effect is less pronounced, with improvements on some benchmarks and minor degradations on others. Notably, the magnitude of these degradations is considerably smaller than the observed gains, suggesting that the anchor term acts as a stable complementary signal to the low-rank rationale alignment. Overall, the largest benefits are observed in smaller models, while the effects become more nuanced at larger scales. Importantly, the proposed distillation method remains effective even without the anchor term, indicating that the global low-rank alignment alone captures a substantial portion of the reasoning signal.
\begin{table}[h]
\small
\centering
\caption{Effect of the anchor loss term on accuracy across models.}
\label{tab:anchor_ablation}
\begin{tabular}{llcc}
\toprule
\textbf{Model} & \textbf{Setting} & \textbf{GSM8K} & \textbf{SVAMP} \\
\midrule

\multirow{2}{*}{Qwen2.5-0.5B}
 & No Anchor & 39.8 & 54.7 \\
 & With Anchor    & \textbf{50.0} & \textbf{63.0} \\
\midrule

\multirow{2}{*}{LLaMA-3.2-1B}
 & No Anchor & 52.8 & \textbf{64.3} \\
 & With Anchor    & \textbf{54.9} & 64.0 \\
\midrule

\multirow{2}{*}{LLaMA-3.2-3B}
 & No Anchor & \textbf{71.3} & 77.3 \\
 & With Anchor    & 70 & \textbf{83.3} \\

\bottomrule
\end{tabular}
\end{table}

\begin{table}[h]
\small
\centering
\caption{Accuracy (\%) of LLaMA-3.2-1B with and without chain-of-thought across different domains, before and after supervised fine-tuning.}
\label{tab:cot_scope}
\setlength{\tabcolsep}{4pt}
\begin{tabular}{lcccc}
\toprule
Method & GSM8K & KodCode & ScienceQA & CSQA \\
\midrule
Base NoCoT & 4.9 & \textbf{23.8} & 62.6 & \textbf{56.9} \\
Base CoT   & \textbf{34.8} & 15.0 & \textbf{66.9} & 49.1 \\
\midrule
NoCoT-SFT  & 30.9 & \textbf{41.2} & \textbf{91.1} & \textbf{90.7} \\
CoT-SFT    & \textbf{58.4} & 32.5 & 86.3 & 74.3 \\
\bottomrule
\end{tabular}
\end{table}

\section{Extension Beyond Mathematical Reasoning}
\label{app:extend_beyond_math}
The effectiveness of LoRi depends on settings where CoT itself provides a meaningful advantage. To further examine the scope of CoT across task domains, we compare standard SFT and CoT-SFT on math, coding, science, and common sense tasks using LLaMA-3.2-1B. As shown in Table~\ref{tab:cot_scope}, CoT-SFT substantially improves performance on GSM8K, while providing little or no advantage over standard SFT on KodCode, ScienceQA~\citep{scienceQA}, and CommonsenseQA (CSQA)~\citep{talmor2019commonsenseqa}. This trend is consistent with prior findings that CoT is most effective for mathematical and symbolic reasoning~\citep{sprague2024cot_or_not}. These results further motivate our primary focus on mathematical reasoning, where CoT provides a meaningful advantage to distill.

\section{Extension to Reasoning-Oriented Models}
\label{app:qwen3}
To study LoRi's performance for a reasoning-oriented model, we additionally evaluate it on Qwen3-0.6B. As seen in Table~\ref{tab:qwen3}, LoRi follows the same overall pattern observed with the non-thinking models: its latent reasoning recovers most of the explicit-CoT accuracy relative to the no-reasoning baseline. In particular, LoRi achieves 63.5\% on GSM8K, 74.0\% on SVAMP, and 24.8\% on GSM8K-Hard, compared with 67.3\%, 73.0\%, and 30.3\% for CoT-SFT.

\begin{table}[h]
\small
\centering
\caption{Accuracy (\%) of LoRi on Qwen3-0.6B.}
\label{tab:qwen3}
\begin{tabular}{lccc}
\toprule
Method & GSM8K & SVAMP & GSM8K-Hard \\
\midrule
NoCoT-SFT & 25.1 & 23.5 & 5.6 \\
CoT-SFT   & \textbf{67.3} & 73.0 & \textbf{30.3} \\
LoRi      & 63.5 & \textbf{74.0} & 24.8 \\
\bottomrule
\end{tabular}
\end{table}

\section{Implementation Details and Hyperparameter}
\label{app:hyperparams}

\begin{table}[h]
\centering
\caption{Hyperparameter settings.}
\label{tab:hyperparameters}
\begin{tabular}{lc}
\toprule
\textbf{Hyperparameter} & \textbf{Value} \\
\midrule
$R_L$ & 8\\
$R_H$ & 256\\
$R_L^{(a)}$ & 8\\
$R_H^{(a)}$ & 128\\
$R_B^{(a)}$ & 64\\
$\lambda$ & 1\\
$\beta$ & $10^{-5}$ \\
$\omega_C$ & 1\\
$\omega_\mu$ & $10^{-2}$\\
Learning rate &  $2 \cdot 10^{-5}$\\
Epochs & 1\\
Samples & 128\\
\bottomrule
\end{tabular}
\end{table}

We assume a pretrained teacher model that produces high-quality
chain-of-thought rationales. The teacher is kept fixed and used to extract
hidden states for both the rationale tokens and the answer prediction
position. From these, we precompute the low-rank projection factors and
anchor-level targets as described in Sec.~\ref{sec:rational-align} and Sec.~\ref{sec:anchor-align}. The student model is then trained using the
proposed loss in Eq.~\eqref{eq:loss}. 

We evaluate the model on standard mathematical reasoning benchmarks,
including GSM8K~\citep{cobbe2021gsm8k}, GSM8K-Hard~\citep{gao2022pal},
and SVAMP~\citep{patel2021svamp}. Hyperparameter settings for LoRi corresponding to the results in Table~\ref{tab:reasoning_results} is reported in Table~\ref{tab:hyperparameters}. Here, $R_L$ and $R_H$ denote the ranks used for the low-rank decomposition of rationale representations, while $R_L^{(a)}$, $R_H^{(a)}$, and $R_B^{(a)}$ are the Tucker ranks for anchor factorization. We use a shared hyperparameter configuration across all models, except for $R_L^{(a)}$, which is increased for larger models (e.g., LLaMA 3B and 8B) to 16 due to their greater depth, while all other parameters remain fixed.

\section{Tensor Contraction}
\label{app:tensor_contraction}

The operation $\times_n$ denotes contraction of a tensor with a matrix along mode $n$. 
In index form, the expression
\[
\ten{X}=\ten{Q} \times_1 \mat{U}_N \times_2 \mat{U}_L \times_3 \mat{U}_H
\]
corresponds to
\begin{align}
X_{n,t,h}
&=
\sum_{m=1}^{r_N}
\sum_{i=1}^{r_L}
\sum_{j=1}^{r_H}
Q_{m,i,j}\,
(\mat{U}_N)_{n,m}
\nonumber\\
&\quad \cdot
(\mat{U}_L)_{t,i}\,
(\mat{U}_H)_{h,j}
\end{align}
which contracts $\ten{X}$ along the first, second and third modes.

\end{document}